\documentclass[english]{pfia}

\usepackage{xcolor}
\usepackage{soul} 

\definecolor{afiablue}{RGB}{61,159,207}
\definecolor{afiared}{RGB}{167,75,68}
\definecolor{afialightblue}{RGB}{158,193,232}

\newcommand{\field}[1]{\textcolor{afiablue}{#1}}
\newcommand{\val}[1]{\textsf{\textcolor{afialightblue}{#1}}}

\newcommand{\LM}[1]{\textcolor{black}{#1}}
\newcommand{\MM}[1]{\textcolor{black}{#1}}

\theoremstyle{definition} 
\newtheorem{definition}{Definition}[section] 

\title{\textbf{A Survey of Multi-Agent Deep Reinforcement Learning\\
               with Graph Neural Network-Based Communication}\footnote{We gratefully acknowledge Université Lyon 1 and the AAP AEC for their support of this research}}

\author{Valentin Cuzin-Rambaud\fup{1}, Laetitia Matignon\fup{1}, Maxime Morge\fup{1} \\[6pt]
\fup{1}Université Lyon 1, INSA Lyon, CNRS, LIRIS, UMR 5205, Lyon, France
}

\date{\{valentin.cuzin-rambaud, laetitia.matignon, maxime.morge\}@univ-lyon1.fr}

\begin{document}

\maketitle


\begin{resume}
\LM{En} apprentissage par renforcement multi-agents (MARL), \LM{l'intégration de} mécanismes de communication permet aux \LM{agents} d'apprendre à coordonner leurs actions et à converger vers leurs objectifs en partageant des informations. Sur la base d'un graphe d'interaction, une sous-classe de méthodes utilise des réseaux de neurones de graphes (GNN) pour apprendre à communiquer, ce qui permet aux agents d'améliorer \LM{leur représentation interne} enrichie par les informations échangées. \LM{Avec l'essor récent} des travaux dans ce domaine, nous constatons un manque de visibilité dans la distinction et la classification des approches MARL \LM{avec communication basée sur les GNNs}. \LM{Ainsi}, cet article passe en revue les travaux récents dans ce domaine. Nous proposons ici un processus généralisé de communication basé sur les GNNs dans le but de rendre plus évidents et accessibles les concepts sous-jacents à ces approches.
\end{resume}

\begin{motscles}
Apprentissage par renforcement multi-agents, Communication, Réseau de neurones de graphes.
\end{motscles}

\begin{abstract}
In multi-agent reinforcement learning (MARL), the integration of a communication mechanism, allowing agents to better learn to coordinate their actions and converge on their objectives by sharing information. Based on an interaction graph, a subclass of methods employs graph neural networks (GNNs) to learn the communication, enabling agents to improve their internal representations by enriching them with information exchanged. With growing research, we note a lack of explicit structure and framework to distinguish and classify MARL approaches with communication based on GNNs. Thus, this paper surveys recent works in this field. We propose a generalized GNN-based communication process with the goal of making the underlying concepts behind the methods more obvious and accessible.
\end{abstract}


\section{Introduction}


\MM{Multi-Agent Systems (MAS) address a wide range of applications in robotics, video games, and cybersecurity. The main difficulty in designing MAS lies in developing the internal mechanisms that govern agents' behaviors and interactions. Reinforcement Learning (RL) provides a framework through which an agent can learn to behave effectively through experience. In the Multi-Agent Reinforcement Learning (MARL) setting, agents learn simultaneously while attempting to coordinate with one another by executing local policies. However, each agent typically operates under partial observability of the global state of the system and must therefore act based on incomplete information. Moreover, because all agents update their policies concurrently, the environment becomes non-stationary \LM{from the point of view of each agent}: the transition dynamics depend on the evolving behaviors of the other agents. This non-stationarity significantly complicates learning in MAS and may prevent convergence to optimal policies.}

To tackle partial observability and non-stationarity in MARL, we can consider communication between agents. Indeed, agents can communicate \LM{some information, e.g. their local observation or an internal representation of their mental state,} to obtain a broader view of the environment and make a well-informed decision. \LM{Yet} communication raises \LM{additional} challenges. \LM{It requires specifying how messages are created, when and with whom communication should occur, how incoming messages are interpreted and aggregated, and how communication is integrated into the learning and execution phases.}

Various approaches have been proposed in the literature to integrate communication into MARL and address its challenges.
\LM{A particularly active research direction in recent years has focused on the use of Graph Neural Networks (GNNs) \cite{scarselli2008graph} for communication in MARL.}
GNNs have emerged as a popular solution to learn and extract knowledge from a graph. When data contains relationships between entities, modeling a graph permits extracting useful information. In many cases, graphs are used to represent data such as infrastructure, biological, social, and collaboration networks. In MARL with a communication context, we can thus represent the state of an environment as a graph where agents are positioned as nodes, and communication between them is represented by edges. Zhu et al. cover MARL with communication \cite{zhu2024survey} more broadly than our work, which focuses on a deeper analysis of twelve major recent GNN-based communication methods, reflecting the growing interest in GNNs for communication.

We note a lack of explicit structure and framework to distinguish and classify MARL approaches with communication based on GNNs, as there are many methods, with great diversity. Hence, this survey analyzes different methods of communication \LM{in MARL based on} GNNs. We motivate the interest in using GNNs, and dive into state-of-the-art methods, \LM{by} comparing them, extracting tendencies, and pointing out limitations of such approaches. Our study leads us to derive a generic algorithm \LM{which generalizes GNN-based communication process in MARL.}

We first take a step back in the background Section \ref{sec:Background} to \LM{present} GNNs, RL, MARL, and MARL with communication. Section \ref{sec:main} presents the generic GNN-based communication process \LM{we propose} and a survey of GNN-based communication MARL methods. \LM{This is followed by a specific focus on how realistic communication constraints are taken into account in state-of-the-art approaches.} Finally, we conclude with future research directions in Section \ref{sec:ccl}.
\section{Background} \label{sec:Background}

In this section, we first outline key aspects of GNNs, \LM{as they} are widely used for communication. Secondly, we briefly establish the foundation of RL, then explore MARL, \LM{focusing progressively} on communication in MARL.

\subsection{Graph Neural Networks} \label{sec:GNN}

\LM{Several tasks on graphs, such as node classification, link prediction, and graph classification leverage machine learning methods \cite{ZHOU202057}. A major approach for solving these tasks is the use of deep learning methods, in particular Graph Neural Networks (GNNs) \cite{scarselli2008graph}.}

\begin{definition} \label{def:G}
	A \textbf{graph} at time $t$, denoted $G^t(V,E)$, is defined by a set of $n$ nodes $V$, and a set of edges $E$, where an edge represents the influence of the source over the target, possibly reciprocal. \LM{$\mathcal{N}(i)$ denotes the set of neighbors of node $i$, as $\forall j\in \mathcal{N}(i), \exists (j,i) \in E$ and $N_i=deg(i)=|\mathcal{N}(i)|$.} The graph can potentially admit self-loops such as $\forall i \in V, (i,i)\in E$ and thus $i$ is included in its neighborhood $i \in \mathcal{N}(i)$. We denote $A \in \mathbb{R}^{n \times n}$ the adjacency matrix, with $A_{ji} = 1 \iff (j,i) \in E$, and $X \in \mathbb{R}^{n \times d}$ the node feature matrix, with $X[i] \in \mathbb{R}^{d}$ being the feature vector of the node $i$ and $d$ the feature dimension. If edges also have features, we denote $E_{attr}$ the edge feature matrix, with $E_{attr}[(j,i)]$ abbreviating $e_{j,i}$ referring to the feature vector of the edge $(j,i)$. The graph is directed and dynamic as it evolves in a finite number of steps $t\in[0,T]$, and can be potentially weighted, with each edge having $ew_{j,i}$ the weight of the edge $(j,i)\in E$.
\end{definition}

\LM{A GNN is defined as a parameterized function $f(X,A,E_{attr};\theta)$.
In a layered GNN with $L$ layers, learnable parameters $\theta$ are a set of weight matrices $\{W^{(l)}\}_{l=0}^{L-1}$ shared among all nodes of the graph. The GNN learns to compute node representations (or embeddings), noted $h_i^{(l)}$ for the representation of the node $i$ at layer $l$. It does so by applying iteratively ($L$ times) the layer process. One layer process consists of two main steps for each node: 1) the \textbf{aggregation} step to gather representations from its neighbors, defined by the adjacency matrix; 2) the \textbf{transformation} step to transform the aggregated information using a weight matrix to obtain updated node representations.
Thus, the edges of the graph determine which neighboring representations are aggregated, while the weight matrix controls how this information is transformed.}

Inspired by the Convolutional Neural Network (CNN), \LM{one of the first GNN architectures is the Graph Convolutional Networks (GCN) \cite{kipf2016semi},} which is applied on non-euclidean data. Applying a convolution on an image uses a kernel with a fixed size, which is impossible in a graph, as the number of neighbors $N_i$ of each node $i$ varies. \LM{Thus, the aggregation step of the GCN layer process consists of pooling neighbors' features, and the transformation step consists of computing the weighted sum of neighbors' values, with learnable weights. The updated representation of a node $i$ at layer $l$ is:}
\begin{equation}
	h_i^{(l)}=\sigma(\sum_{j\in \mathcal{N}(i)}\frac{ew_{j,i}}{\sqrt{deg(i)deg(j)}}h_j^{(l-1)}W^{(l-1)})
	\label{eq:GCN}
\end{equation}
The average is weighted by the degree of nodes preventing high-degree nodes from dominating the aggregation and keeping the feature magnitudes stable.

More recently, Graph Attention Networks (GAT) \cite{velivckovic2017graph} proposes a new model based on attention heads:
\begin{equation}
	h_i^{(l)}=\sum_{j\in \mathcal{N}(i)}\alpha_{ji}h_j^{(l-1)}W^{(l-1)}
	\label{eq:GAT}
\end{equation}
with $\alpha_{ji}$ the attention weights from $j$ to $i$. The use of a learnable attention vector allows choosing how much attention to give to each neighbor.

GCN and GAT are two popular architectures \LM{among the many variants of GNNs. Most of them} can be instantiated to a single common framework: Message Passing Neural Networks (MPNNs) \cite{gilmer2017neural}. \LM{The layer process with MPNN is}:
\begin{equation}
	h_i^{(l)}=\psi^{(l)}(h_i^{(l-1)}, \bigoplus_{j\in \mathcal{N}(i)} \phi^{(l)}(h_i^{(l-1)},h_j^{(l-1)},e_{j,i}))
	\label{eq:MPNN}
\end{equation}
with $\psi$ and $\phi$ differentiable functions, e.g. Multi Layer Perceptrons (MLP), and $\bigoplus$ the aggregator function, (e.g. mean, add, max). $\phi^{(l)}$ represents the message construction, and for GCN and GAT cases, it would contain weights $W^{(l)}$.

\begin{figure}[h!t]
	\centering
	\includegraphics[width=\linewidth]{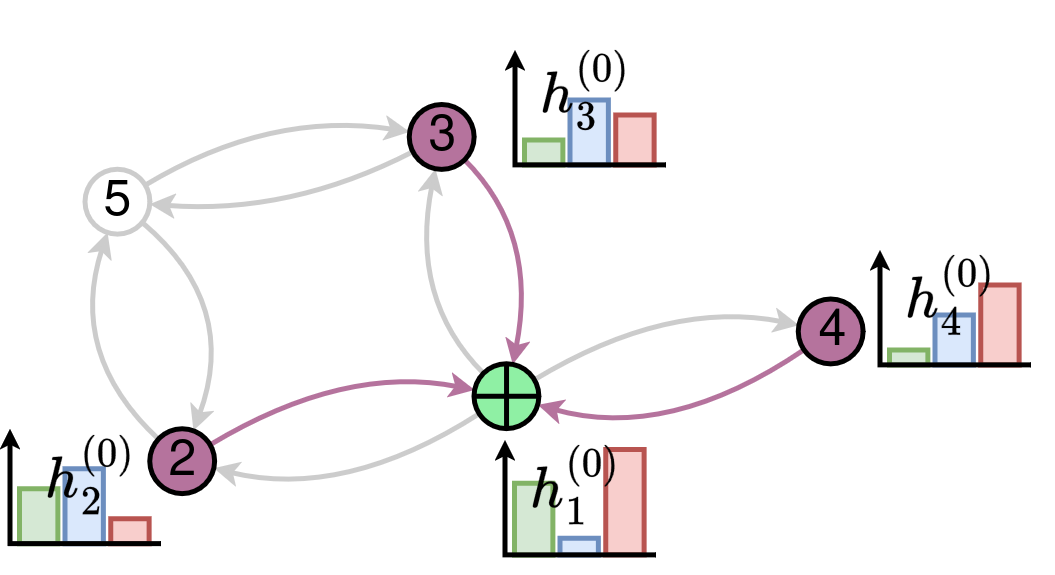}
	\caption{An example of MPNN: the green node, aggregates features from its 1-hop neighbors $\in \mathcal{N}(1)$}
	\label{fig:GNNExplained}
\end{figure}

The main advantage of using MPNN is transmitting information in the graphs to $l$-hops by stacking layers, with nodes communicating only with immediate neighbors. In Figure \ref{fig:GNNExplained}, the green node aggregates information from its neighbors (nodes 2,3 and 4) \LM{in 1-hop. Repeating the process} permits accessing information from 2-hop nodes (node 5) as their features were previously aggregated by 1-hop nodes. Other advantages of GNNs include generalizability to unseen nodes (inductive learning), invariance of permuting node order during aggregation, handling of non-Euclidean structure, and capturing local and global information to learn complex patterns. All these advantages have led communication methods to integrate GNNs in their process. In particular, in distributed communication using GNNs with $l>1$, agents \LM{(considered as nodes)} indirectly aggregate information about unreachable agents, \MM{effectively extending information propagation despite limited communication range.}

\subsection{Reinforcement Learning} \label{sec:RL}
\LM{In a sequential decision-making setting, reinforcement learning (RL) algorithms enable an agent to learn solutions through repeated experiments with an environment.}

The standard model to define the sequential decision process is the Markov Decision Process (MDP) defined with $S$ the set of environment states and $\mu$ its initial state distribution such that $s_0 \sim \mu$, $A$ the set of actions for the agent, $\mathcal{R}$ the reward function and $\mathcal{T}$ the state transition probability function. In a more realistic case, the agent only partially observes the state of the environment. Formally, we define the Partially Observable Markov Decision Process (POMDP) as an MDP extended with $O$, the set of observations. $\mathcal{O}$ \LM{defines a probability distribution over possible observations.}
The agent interacts with the environment during \LM{a set of} episodes (i.e. sequence of states and actions until a terminal state).

The discounted return \LM{from time step $t$ is defined as the sum of discounted rewards over time, i.e. $R_t=\sum_{k=0}^{\infty}\gamma^k r_{t+k}$} where the discount factor $\gamma \in [0,1[$ ensures finite return in non-terminating MDPs, \LM{and $r_t$ is the reward received at time step $t$}. \LM{A policy $\pi(a|s)$ defines a mapping from states to a probability distribution over actions.} The goal of an agent is to find an optimal policy $\pi^*$ which maximizes the \textit{expected discounted return} \LM{from every state $s\in S$, i.e. $\pi^* = \arg\max_\pi \mathbb{E}_\pi[R_t | s_t=s]$.}
We can define two useful functions: the value function $V^\pi(s)=\mathbb{E}_\pi[R_t | s_t=s]$, and the action-value function $Q^\pi(s,a)=\mathbb{E}_\pi[R_t | s_t=s,a_t=a]$ which enable to evaluate the expected return based on a state $s$ or a couple $(s,a)$. The use of $V$ or $Q$ is essential, as an action can lead to a high immediate reward, but in the long term, a poor return.

In RL, \LM{a common assumption is} that the agent has no a priori knowledge \LM{about the transition and reward functions}, leading to the need to collect experiences to learn a policy $\pi$ \cite{watkins1992q}.

The use of \LM{deep neural networks to parameterize and approximate value functions and policies over large state and action spaces has become standard practice, giving rise to Deep Reinforcement Learning (DRL).} We can classify DRL algorithms into two categories: value-based algorithms, which optimize a value function like DQN \cite{mnih2013playing}; and policy-based algorithms, which optimize the policy, like REINFORCE \cite{williams1992simple}. A particular subclass of policy-based algorithm is the actor-critic algorithm, which combines the use of a value function as a critic to guide policy learning such as A2C \cite{mnih2016asynchronous} or PPO \cite{schulman2017proximal}.



\subsection{Multi-Agent Reinforcement Learning} \label{sec:MARL}

\LM{In a multi-agent setting, a POMDP extends to a Partially Observable Stochastic Game (POSG).}

\begin{definition}
	A \textbf{POSG} \cite{marl-book} is defined by a tuple $\langle I,S,\mu,\{O_i\},\{A_i\},\{\mathcal{R}_i\},\mathcal{T},\{\mathcal{O}_i\}\rangle$, with $I$ the set of agents indexed as $\{1,\dots,n\}$\footnote{In the multi-agent setting, time steps are denoted using superscripts rather than subscripts to avoid confusion with agent indices.}, $S$ the set of states and $\mu$ the initial state distribution such as $s^0 \sim \mu$, $O_i$ the set of observations for agent $i$, $A_i$ the set of actions. We denote the joint action space ${\bf A}= \times_{i \in I}A_i$. Thus, $\mathcal{R}_i:S\times {\bf A} \times S \rightarrow \mathbb{R}$ is the reward function \LM{for agent $i$}, $\mathcal{T}:S\times {\bf A} \times S \rightarrow [0,1]$ is the state transition probability function, and $\mathcal{O}_i:S\times {\bf A}\times O_i \rightarrow [0,1]$ is the observation function of agent $i$.
\end{definition}

At each time step $t$, each agent $i$ receives a local observation of the current state $o_i^t\subset s^t$ given by its observation function $\mathcal{O}_i(o_i^t|s^t,a^{t-1})$. All the individual observations give the joint observation $o^t=\langle o_1^t,\dots,o_n^t \rangle$, which approximates the current state $o_i^t\subset o^t \subseteq s^t$. In addition, each agent can memorize its observation-action history $\tau_i^t=[(o_i^0,a_i^0),\dots,(o_i^t,a_i^t)]$ often encoded by a Recurrent Neural Network (RNN). Thus, by following its policy $\pi_i(a_i^t|\tau_i^t)$, each agent $i$ chooses action $a_i^t$, forming the joint action $a^t=\langle a_1^t,\dots,a_n^t \rangle$. The game transitions to the next state $s^{t+1}\in S$ with $\mathcal{T}(s^{t+1}|s^t,a^t)$ probability, and \LM{each} agent $i$ receives its reward $r_i^t=\mathcal{R}_i(s^t,a^t,s^{t+1})$. We note $R_i=\sum_{t=0}^{\infty}\gamma^t r_i^t$ the discounted return for the agent $i$.

\LM{Depending on the reward structure, POSGs can be classified as competitive, cooperative, or mixed games. In competitive games, agents are opponents (e.g., $\sum_{i \in I}\mathcal{R}_i(a)=0, \forall a \in \bf{A}$).}
In cooperative games, agents aim to maximize the global expected return of all agents $\sum_{i\in I}\mathbb{E}_{\pi_i}[R_i]$ (e.g., common-reward: $\mathcal{R}_1=\mathcal{R}_2=\dots=\mathcal{R}_n$).
\LM{In mixed games, agents may share partially aligned interests while still pursuing individual objectives. }

\paragraph{Example: Predator-Prey.} \label{ex:PP} The Predator-Prey environment is well-known and massively used to evaluate MARL algorithms. Let $n$ \LM{agents (predators) evolve in a square grid by choosing a movement action \{up, down, left, right, stay\}. Their objective is to catch a stationary prey while constrained by a limited vision range. Once a predator reaches the prey, it stays there and always gets a positive reward until the end of the episode (reaching the time step limit or all predators have reached the prey). The environment is cooperative: agents obtain a better reward when more agents reach the prey.}

Multi-Agent DRL (MADRL) algorithms fall into two paradigms.

\paragraph{Decentralized Training and Execution (DTDE).} \LM{DTDE addresses the problem independently for each agent}. Each agent uses its own local information to choose its own behavior. This learning framework permits both training and execution in a physically distributed setting, such as robotic scenarios. All RL algorithms can be adapted to DTDE, e.g. IDQN \cite{tampuu2017multiagent} and IPPO \cite{de2020independent}. However, the main problem resides in the non-stationarity of the environment, as many agents learn \LM{simultaneously}. From the point of view of one agent, all other agents are \LM{part of} the environment, so the environment always changes even without its action. 
\paragraph{Centralized Training and Decentralized Execution (CTDE).}
\MM{In the CTDE setting, additional global information can be centralized and exploited during training to stabilize and improve learning. However, at execution time, each agent follows a decentralized policy using only its own local observations, ensuring fully decentralized execution.}
The use of centralized information during training helps to reduce the non-stationarity environment problem. In most environments, centralized information is accessible only during training, as the training process is typically simulated. This is the reason why many popular algorithms follow the CTDE paradigm. Some use value decomposition to centrally combine individual utility functions during training, like VDN \cite{sunehag2017value} or QMIX \cite{rashid2020monotonic}, while others use a centralized critic and decentralized policies like MADDPG \cite{lowe2017multi} and MAPPO \cite{yu2022surprising}.

Whatever the paradigm is, agents can either share the same set of network parameters or not. If agents share their parameters, only one set of parameters is learned and then duplicated into each agent, for the decentralized execution. In common-reward cooperative games, sharing weights of the model during learning permits converging faster, but at the cost of having less chance to develop diversity in behaviors between policies \cite{marl-book,bettini2023heterogeneous}.

\subsection{MARL with Communication} \label{sec:marl-Comm}

Classical CTDE settings assume no communication between agents during execution. However, the centralization of information during training can be seen as an implicit communication process: all agents send their local observation to a centralized node that aggregates information to form the joint observation. \LM{Indeed, the messages exchanged during the training phase are not formally specified.} In contrast, the explicit communication process that we refer here is learnable, \LM{where information is explicitly communicated between agents}. This grant a broader representation of the global state for each agent, while remaining in a decentralized execution setting.

Most methods use communication during execution to balance the partial observability while enabling better coordination strategies.
The communication process can also be used only during training, for instance \LM{to learn a more effective information combination than predefined CTDE schemes.}
\LM{The positive impact of communication on coordination explains why cooperative scenarios are the most relevant setting for its use.}

\LM{Incorporating explicit communication in MARL requires extending the joint action space to include communication actions.}
Zhu et al. extend POSG with $\mathcal{M}$, the shared message space, leading to POSG-Comm \cite{zhu2024survey}.

Learning to communicate brings some \textbf{new challenges}: 
\begin{enumerate}
	\item Generating the content of the message to be sent.
	\item Choosing when and with whom to communicate.
	\item Interpreting/combining received messages.
	\item \LM{Leveraging the acquired knowledge.}
\end{enumerate}
Many methods have been proposed to solve these new challenges \cite{zhu2024survey}. Each of these issues can be resolved \LM{either through learning or by relying on a fixed, manually designed solution.} \LM{Most state-of-the-art approaches incorporate learning to handle at least one of these challenges}. The communication is often learned in an end-to-end fashion (with backpropagation flow for all differentiable functions), so agents communicate directly to improve the global return.

\LM{A key distinction between MARL methods with communication is whether the communication requires a central control node or not.} This means that some methods need a centralized data structure to control the communication policy. This is endorsed by the proxy, an agent that cannot directly interact with the environment but is responsible for the communication policy, e.g. learning to aggregate observations for a centralized critic. The use of a proxy during the training phase is a \LM{soft} constraint, as it is often implemented within a simulator \LM{and centralized}. Yet during execution it is a harder constraint, as it requires perfect communication between all the agents and the proxy. \LM{While this constraint may be realistic in some settings, such as warehouse environments, it restricts the applicability to other contexts, e.g. fleet of drones.} In methods without a proxy, each agent communicates in a distributed manner, within a communication-range, which makes the approach more flexible and applicable to a wider set of scenarios.

\begin{algorithm}[h!t]
\caption{General Pipeline for MARL with communication}
\label{alg:MARL}
\SetKwComment{Comment}{$\triangleright$\ }{}
\SetKwInput{Input}{Input}
\SetKwInput{Output}{Output}
\Input{The environment $env$}
Initialize all parametric functions $f_i(\cdot;\theta_i)$ with their associated weights $\theta_i$ \;
\For{every episode}{
	Observe at $o_1^0,\dots,o_n^0$ from $env$ at $t=0$ \;

	\For{time step $t=1,2,\dots,T-1$}{

		\Comment{Create representations with Algo. \ref{alg:comm}}
		$h_1^t,\dots,h_n^t = Communicate(o_1^t,\dots,o_n^t)$ \;
		Sample actions $a_1^t,\dots,a_n^t \sim \pi_1^t(\cdot|h_1^t;\theta_1),\dots,\pi_n^t(\cdot|h_n^t;\theta_n)$ \;
		Perform actions in $env$ \; 
		Collect rewards $r_1^t,\dots,r_n^t$ from $env$ \;
		Observe $o_1^{t+1},\dots,o_n^{t+1}$ from $env$ \;
		\If{in Training phase}{
			Compute losses for parametric functions, based on agents' interactions with $env$ \;
			Update $\theta_i$ by backpropagating gradients \;
		}
	}
}
\end{algorithm}

Algorithm \ref{alg:MARL} summarizes the process for the training and the execution phases.  Parametric functions are any common learnable function $f(\cdot;\theta)$ with associated weight $\theta$. For example, it can be used to compute action probabilities for the policy or to serve a critic in actor-critic methods.
The communication process permits obtaining relevant information by exchanging messages between agents, which influences the policy of agents. Agents are either fully connected, or constrained by the communication range. Various mechanisms using GNNs to address communication challenges will be explored in the following section.

\section{MADRL with GNN-Based Communication} \label{sec:main}

GNNs excel at propagating information between nodes \LM{and at modeling structured and dynamic interactions among agents \cite{Jiang2020Graph,li2021deep,liu2020multi}.
In this section, we first abstract and generalize the communication process used in existing state-of-the-art approaches to design a generic} algorithm for communication in MADRL. Secondly, we survey GNN-based communication methods, their advances, and limitations, including the widespread neglect of communication constraints.

\subsection{Communication Model with GNNs} \label{sec:building_comm}
Introducing GNNs in the communication process can help to solve challenges. Creating messages through GNNs permits communication in multiple rounds. Building a graph at each timestep and designing an MPNN determines when\footnote{At each timestep, the agent decides whom to communicate with, so agents implicitly learn when to communicate.} and with whom to communicate, and how to create messages and interpret received messages. The last aggregation and transformation made by the MPNN gives a final representation (i.e., communication-aware representation) that is integrated in the learning process, and optionally in the execution.

To aggregate many GNN-based methods within a high-level framework, we \LM{propose a generic} communication process using GNNs in MADRL, \LM{presented in} Algorithm \ref{alg:comm}. \LM{We distinguish between two cases: the use of a proxy or distributed communication without a proxy.} The associated proxy part \LM{is detailed in} Algorithm \ref{alg:proxy comm}.

\begin{algorithm}[ht]
\caption{Generic communication process in MADRL with GNNs at time $t$ for agent $i$}
\label{alg:comm}
\SetKwComment{Comment}{$\triangleright$\ }{}
\SetKwInput{Input}{Input}
\SetKwInput{Output}{Output}
\Input{$o_i^t$: the local observation}
\BlankLine

\Comment{Encode the message to send}
Encode current representation: $x_i^t\leftarrow E(o_i^t)$ \;
\BlankLine

\Comment{Decide with whom to communicate}
\If{exists a proxy $P$}{
	Send $x_i^t$ to $P$ \;
	\Comment{Execute Algo. \ref{alg:proxy comm}}
	Receive $h_{P,i}^{t,L}$ from $P$\;
}
\If{exists a distributed comm}{
	Send $x_i^t$ to $\{ j \in \mathcal{N}^t_r(i) \}$ \;
	$X_i^t\leftarrow\{ x_j^t | i \in \mathcal{N}^t_r(j), \forall j \in I\}$ \;
	\BlankLine

	\Comment{Combine received msg}
	$G_i^t(V_i^t,E_i^t)\leftarrow G_{build}(X_i^t)$: the local graph \;
	$H_i^{t,0}=X_i^t$ \;
	\For{$l=1\dots L$}{
		\Comment{Aggregate msg}
		$h_i^{t,l} \leftarrow GNN_i^l(H_i^{t,l-1}, E_i^t)[i]$ \;
		\If{multi-round comm. and $l > 1$}{
			\Comment{Exchange updated representation}
			Send $h_i^{t,l}$ to $\{ j \in \mathcal{N}^t_r(i) \}$ \;
			$H_i^{t,l}\leftarrow\{ h_j^{t,l} | i \in \mathcal{N}^t_r(j), \forall j \in I\}$ \;
		}
		\If{evolution of relation}{
			$G_i^t(V_i^t,E_i^t)\leftarrow G_{build}^l(H_i^{t,l})$ \;
		}
	}
	\BlankLine
}
\Output{The final representation $h_i^{t,L}$, Possibly final representation from the proxy $h_{P,i}^{t,L}$}
\Comment{Inner integration to policy/value-level}
\end{algorithm}

The algorithm can be unfolded into four main parts:

\paragraph{1. Encode the message to send.} To generate the first message to send at time $t$, we use $E(\cdot)$, the message encoder: it maps local observation of the agent $o_i^t$ to encoded message $x_i^t$ (\LM{line 1}). It can be implemented as any parametric function (e.g., MLP, RNN, CNN), or as a predefined mapping, such as the identity function, which preserves the raw input. Using an RNN introduces a hidden state to capture history.

\paragraph{2. Decide with whom to communicate.} In proxy communication settings, all agents send their local representations to the proxy (\LM{line 3}). In distributed settings, each agent $i$ sends its representation to agents $j\in \mathcal{N}^t_r(i)$ with $\mathcal{N}^t_r(i)$ containing the set of reachable neighbors from $i$, at time $t$ (\LM{line 6}). Each agent $i$, received messages from every agent $j\in I$, that considers $i\in \mathcal{N}^t_r(j)$ (line 7). $\mathcal{N}^t_r(k)$ is determined by potential communication range or learnable choices to \LM{decide with whom it is possible or beneficial to communicate}.

\paragraph{3a. Combine received messages with distributed communication.} \LM{Following the previous message-sending phase at time $t$, each agent $i$ receives a set of messages $X_i^t$ from the others (line 7).}
A local graph for agent $i$ at time $t$, noted $G_i^t$, is constructed by $G_{build}(\cdot)$ function, with all previous received messages $X_i^t$ (line 8). This local graph $G_i^t$ represents all reachable agents from $i$ (all $j \in \mathcal{N}^t_r(i))$ as nodes, their communication as edges, and message contents as node feature $X_i^t[j]: \forall j\in \mathcal{N}^t_r(i)$ (see Def. \ref{def:G}). $G_i^t$ is then used to aggregate messages with GNNs.
Combining messages until the last layer $L$ of GNNs leads the agent to obtain a final representation \LM{for its corresponding node in the graph}, $h_i^{t,L}$, which is intended to be a broader representation of the current global state of the world.
If $l=1$, information is aggregated only with the 1-hop neighbor, but if $l>1$, two possible cases occur. In the \textit{multi-round communication} case, at each layer of GNNs, each agent communicates with its neighbors, and its intern representation $h^{t,l}_i$ is updated at layer $l$ (line 13). The \textit{multi-round communication} leverages the diffusion/propagation mechanism of MPNN (see Fig. \ref{fig:GNNExplained}).
Otherwise, each agent $i$ computes representations $l$ times, for all reachable nodes, which implies that the $i$'s representation of agent $j$ may differ from $j$'s self-representation (i.e., they do not have the same local graph).
The multi-round communication ensures construction of more globally consistent representations, but demands more messages. Whereas without multi-round communication, agents communicate only once, but representations have limited global consistency and may be biased. Furthermore, with the \textit{evolution of relation} enabled, $G_i^t$ can be updated at each layer via $G_{build}^l(\cdot)$ to learn a different communication structure within the same timestep (line 16).
The final representation $h_i^{t,L}$ is integrated in the MADRL pipeline in the next phase. The final representation replaces the usage of raw observation $o_i^t$.

\begin{algorithm}[!htbp]
	\caption{Proxy' communication process at time $t$}
	\label{alg:proxy comm}
	Receive $X_P^t \leftarrow \{x_i^t|\forall i \in I\}$ \;
	$G_P^t(V_P^t,E_P^t)\leftarrow {G_{build}}(X_P^t)$: the graph of Proxy \;
	$H_P^{t,0}=X_P^t$ \;
	\For{$l=1\dots L$}{
		$H_P^{t,l} \leftarrow GNN_P^l(H_P^{t,l-1}, E_P^t)$ \;
		\If{evolution of relation}{
			$G_P^t(V_P^t,E_P^t)\leftarrow {G_{build}}^l(H_P^{t,l})$ \;
		}
	}
	$\forall i \in I,$ send $H_P^{t,L}[i]$ \;
\end{algorithm}

\paragraph{3b. Combine received messages with the proxy (Algo. \ref{alg:proxy comm}).} In proxy communication, the construction of the graph is centralized. \LM{The proxy receives messages from all agents $X_P^t$ (line 1)}. This global view allows to build a more globally consistent graph \LM{than the distributed communication}, where all nodes are represented, and edges are not restricted by a communication range (line 2). Communication is established once, with the \LM{strong} requirement that all agents are connected to the proxy. \LM{Information of nodes are aggregated through successive layers via the GNN}. The final \LM{joint} representation \LM{of all nodes} $H_P^{t,L}$ can be used directly as centralized information (replacing the joint observation), and \LM{the proxy} sends back to each agent its self-representation $h_{P,i}^{t,L}$\LM{$=H_P^{t,L}[i] $}.

\paragraph{4. Inner integration to policy/value-level.}
The \LM{obtained} final representation $h_i^{t,L}$ is leveraged in training, execution, or both, depending on the MADRL method (value-based, value-decomposition, or actor-critic). If a proxy exists, each agent retrieves its final representation made by the proxy $h_{P,i}^{t,L}$ and can potentially use its. We note that the full representation $H^{t,L}$ can serve for centralized training. The final representation can be combined with other knowledge of the agent before any usage, e.g. the concatenation with raw observation $h_i^{t,L}\leftarrow[o_i^t \oplus h_i^{t,L}]$. The communication is often learned in an end-to-end manner, but in specific methods, an additional objective function can update communication-dependent weights during supervised, self-supervised, or reinforcement learning.

\paragraph{Example: Predator-Prey with distributed communication.} We assume that three agents are in range at time $t$. If agents communicate their local observations, they can cover a broad search space. \MM{By sharing information about previously explored areas where the prey was not observed, agents can take more informed actions.} If agent $i$ finds the prey, its message will inform its two neighbors. With \textit{multi-round communication}, the information can propagate beyond the communication range limit of the agent, similarly to Figure \ref{fig:GNNExplained}. Any agent connected to the real communication graph, and at $L-hop$ from agent $i$, can obtain the prey's position.

\subsection{GNN for communication} \label{sec:GNN4comm}
\LM{Building upon the generic algorithm introduced earlier, we provide in this section a survey of GNN-based communication MADRL methods by instantiating each component of our generic algorithm (cf. Table \ref{tab:gnn4comm}).}
\MM{The proposed generic algorithm facilitates comparative analysis, reveals common tendencies and structural patterns, and enables the classification of existing methods.}

\begin{table*}[h!t]
	\centering
	\caption{\LM{Taxonomy of GNN-based communication methods, organized in two categories: proxy-based (upper part) and distributed-based (lower part) according to the generic Algorithm \ref{alg:comm}.} $E(\cdot)$ encodes the local representation, $\mathcal{N}^t_r(\cdot)$ defines reachable agents, $G_{build}(\cdot)$ builds a graph, $GNN$ is the architecture used, and Inner integration shows how to leverage the final representation.} 
	\label{tab:gnn4comm}
	\resizebox{\textwidth}{!}{
	\rowcolors{2}{gray!15}{white}
	\begin{tabular}{lccccc}
		\toprule
		Methods & $E(\cdot)$ & $\mathcal{N}^t_r(\cdot)$ & $G_{build}(\cdot)$ & $GNN$ & Inner integration \\
		\midrule
		DICG \cite{li2021deep} & $\{LSTM,MLP\}(o_i^t)$ & All agents & $G_P^t:$ complete, weighted & $L=2$, $GCN$ & $V(H_P^{t,L})$ \\
		GA-Comm \cite{liu2020multi} & $LSTM(MLP(o_i^t))$ & All agents & $G_P^t:$ sparse, weighted & $L=1$, $MPNN$ with attention & $\pi_i(\cdot|h_{P,i}^{t,L})$ \\
		GAAC \cite{liu2020multi} & $LSTM(MLP(o_i^t))$ & All agents & $G_P^t:$ sparse, weighted & $L=1$, $MPNN$ with attention & $Q_i(h_{P,i}^{t,L})$ \\
		MAGIC \cite{niu2021multi} & $MLP(LSTM(MLP(o_i^t)))$ & All agents & $G_P^{t,l}:$ sparse, dynamic & $L=2/3$, $GAT$ & $\pi_i(\cdot|h_{P,i}^{t,L}), V_i(h_{P,i}^{t,L})$ \\
		GACG \cite{duan2024group} & $MLP(o_i^t)$ & All agents & $G_P^t:$ sparse, weighted, group & $L=2$, $GCN$ & $Q_i(h_{P,i}^{t,L})$ \\
		LTS-CG \cite{duan2024inferring} & $MLP(o_i^t)$ & All agents & $G_P^t:$ sparse, weighted, temporal learned & $L=2$, $GCN$ & $Q_i(h_{P,i}^{t,L})$ \\
		\midrule
		DGN \cite{Jiang2020Graph} & $MLP(o_i^t)$ & near agents & $G_i^t:$ sparse & $L=2$, $MPNN$ with attention & $Q_i(h_i^{t,L})$ \\
		LSC \cite{liu2023deep} & $MLP(o_i^t)$ & near hierarchic agents & $G_i^t:$ sparse, hierarchic, heterogeneous & $L=3$, $MPNN$ & $Q_i(h_i^{t,L})$ \\
		MAGE-X \cite{yang2023learning} & $MLP(o_i^t)$ & near agents & $G_i^t:$ sparse & $L=2$, $GCN$ & $\pi_i(\cdot|h_i^{t,L})$ \\
		MAGEC \cite{goeckner2024graph} & $Identity(o_i^t)$ & near agents & $G_i^t:$ sparse, heterogeneous & $L=10$, $GraphSAGE$ & $\pi_i(\cdot|h_i^{t,L})$, $V_i(h_i^{t,L})$ \\
		(Het)GPPO \cite{bettini2023heterogeneous} & $Identity(o_i^t)$ & near agents & $G_i^t:$ sparse, undirected & $L=1$, $MPNN$ & $\pi_i(\cdot|h_i^{t,L}),  V_i(h_i^{t,L})$  \\
		HetNet \cite{seraj2022learning} & $MLP_i(LSTM(MLP(o_i^t)))$ & near agents & $G_i^t:$ sparse, undirected & $L=3$, $HetGAT$ & $\pi_i(\cdot|h_i^{t,L}), V(H_{P,i}^{t,L})$ \\
		\bottomrule
	\end{tabular}
	}
\end{table*}

\paragraph{Proxy-Communication Methods.} Many methods are based on a proxy to handle all communications. One of the first methods to use a proxy is Deep Implicit Coordination Graphs (DICG) \cite{li2021deep}, which extends MAPPO \cite{yu2022surprising} and uses communication for the centralized critic only. This permits fully decentralized execution since the critic is used only during training.
DICG uses the proxy to compute a complete weighted graph called Implicit Coordination Graph. Each edge obtains a weight computed by a self-attention mechanism. The proxy then applies a GCN (Eq. \ref{eq:GCN}) on the entire graph to obtain the final \LM{joint} representation matrix, which feeds the centralized critic. Basically, DICG aggregates observations instead of using a concatenation of local observations to create the joint observation. 

Unlike DICG, Game Abstraction Communication (GA-Comm) \cite{liu2020multi} uses a proxy for both training and execution. The graph is built in two phases: hard attention (dropping edge) and soft attention (weighting all remain edges via self-attention). This process is called the game abstraction of agents' interactions.
The GNN used is a custom one (Eq. \ref{eq:MPNN}), resulting from aggregation weighted by both hard and soft attention computed weights: $h_i=\sum_{j\neq i}W_h^{i,j}W_s^{i,j}h_j$. The final representation is then used for the policy trained with an independent multi-agent version of the REINFORCE algorithm \cite{williams1992simple}. 
A second implementation called Game Abstraction Actor-Critic (GAAC) is used to compute the local $Q_i$ critic, thus, execution remains totally decentralized, without communication \cite{liu2020multi}.

Multi-Agent Graph-attention Communication (MAGIC) \cite{niu2021multi} can be seen as an upgrade of GA-Comm, as it keeps the principle of hard and soft attention. The hard attention is handled here by a “sub-scheduler“ process, while soft attention is directly learn through GAT (Eq. \ref{eq:GAT}). The first sub-scheduler uses a GAT in the complete graph to compute the first node feature matrix $X_P^t$, then it processes through a sample of effective edges (probabilities to binary). Any additional sub-scheduler reuses $X_P^t$ and samples a new adjacency matrix, updating the graph. The key contribution is the possibility of stacking several sub-schedulers followed by GAT aggregation. Thus, the graph is rebuilt between each aggregation of messages, (cf. line 6 of Algo. \ref{alg:proxy comm}). Several layers of GNNs with evolving structure permit learning different kinds of relationships at each timestep. All created representations of an agent are concatenated and used for critic during training, and for policy during both training and execution.


Group Aware Coordination Graph (GACG) \cite{duan2024group} builds a graph by enforcing the group structure to help agents cohesion. First, groups are formed with a classifier looking at a temporal window in the joint-observation history: if two agents have similar representations, they are in the same group. Then, the proxy computes the agent-pair matrix, which encode weight of relation between all agents, and besides it computes the edge-group matrix, which encodes if edges belong or not to the same group. A weighted adjacency matrix is then sample from a Gaussian distribution based on the agent-pair matrix as mean and the edge-group matrix as covariance. The method, which is built on top of QMIX \cite{rashid2020monotonic}, uses a GCN to aggregate messages, and the final representation is integrated into individual $Q_i$ values. 

Latent Temporal Sparse Coordination Graph (LTS-CG) \cite{duan2024inferring} is another proxy method, focusing on building graphs by leveraging temporal information. First, the method creates an agent-pair matrix as GACG, from correlations between previous trajectories of agents. Then, a graph is sampled on a Bernoulli distribution with probabilities from the agent-group matrix.
The key contribution comes from two pretext tasks used during the learning: predict-future observation and infer-present states. These tasks guide graph construction toward a latent temporal sparse graph that integrates temporal information in the structure of the graph. Furthermore, the graph is weighted and uses GCN (Eq. \ref{eq:GCN}) to aggregate messages, integrated in the individual $Q_i$ value for the QMIX framework \cite{rashid2020monotonic}. 


Using a proxy during the execution requires all agents to have perfect communication with the proxy. Even \LM{under} decentralized execution, the proxy centralizes the communication policy. So the proxy is a very strong constraint, and proxy-methods can not adapt to fully decentralized execution. It is worth noting that \LM{among proxy-based methods, only} DICG and GAAC assume a fully decentralized execution, as they use the proxy only during training and do not communicate during execution.

\paragraph{Distributed Communication Methods.} \LM{As discussed previously}, using a proxy is a strong constraint on the environment, and establishing communication in a distributed manner among agents enables a wider range of applications. Moreover, communications between nearby agents are often more relevant than those between distant agents, because distant agents' observations are often not useful to others and can introduce irrelevant context.
For this purpose, Graph Convolutional Reinforcement Learning (DGN) extends IDQN with GNNs for communication \cite{Jiang2020Graph}. Each agent exchange its \LM{encoded message} to reachable neighbors (cf. lines 6-7 of Algo. \ref{alg:comm}). A custom MPNN (Eq. \ref{eq:MPNN}) aggregates messages using self-attention weights during averaging. DGN enable \textit{multi-round communication} (cf. lines 12-14 of Algo. \ref{alg:comm}), so \LM{each agent re-sends its updated representation $h_i^{t,1}$ to its neighbors}.
As explained in Example \ref{ex:PP}, this leverages the MPNN propagation mechanism (as in Fig. \ref{fig:GNNExplained}).
The inner integration plays a role in training and execution as $Q_i$ is conditioned on the concatenation of $h_i^{t,1}$ and $h_i^{t,2}$ for helping the model to understand different kinds of relationships. 

Scaling to larger number of agents is challenging in MARL. Using a hierarchical structure leverage more sparse communication topology. Learning Structured Communication (LSC) exploit a two-level hierarchical structure, with two agent types: low and high \cite{sheng2022learning}. All low agents can communicate only with high-level neighbors. High agents are elected depending on a computed weight attribute based on local observations. If in the agent $i$ range there is no high agent, and the $i$'s weight is bigger than neighbors' weight, $i$ becomes the new high agent. Agents do not aggregate information in the same ways, and thus, the behavior is not the same for all agents. Integrating \LM{this approach} into our generic algorithm \LM{implies that} high-level agent receive information, build a local graph $G_i$, and aggregate information at $l=1$ with a GNN. Then, they exchange with high-level agents only, updating topology of $G_i$ (cf. lines 15-16 of Algo. \ref{alg:comm}), and perform a second aggregation at $l=2$. Finally, they send back final representations to each reachable low-level agent. A low-level agent, just sends information to high-level agents and receives an answer from them, which will be aggregated only once with $L_{low}=1$. Their custom GNN sums received features and updates via a MLP (like a GCN without degree-normalization). The inner integration is on computing local $Q_i$ values, during both training and execution, as LSC extends IDQN \cite{tampuu2017multiagent}.

\MM{Although Multi-Agent Graph-Enhanced Commander-Executor (MAGE-X) \cite{yang2023learning} is presented as a distributed communication method, it still requires centralization at execution time. A supervising agent first assigns a goal to each agent in navigation tasks. Then, each agent constructs a local complete subgraph that includes all reachable neighbors. A first GCN is applied to this subgraph to compute node attributes, after which an adjacency sampling process is performed to obtain the graph $G_t^i$. A second GCN is then applied to $G_t^i$ to produce the final representation. This representation is combined with the encoded goal assignment and integrated into both the policy and value functions within an IPPO-based framework \cite{de2020independent}.}

Multi-Agent Graph Embedding-based Coordination (MAGEC) \cite{goeckner2024graph} addresses the patrolling problem, where agents navigate a graph to minimize the node idleness. This method builds a heterogeneous graph that includes both game's nodes and neighboring agents. An agent builds first a sub-graph with its limited observation: each node contains the type of node (agent/game node), the idleness time, and the degree of the node. Edges include a relative-position attribute. Agents communicate their position, goal-reached notifications, and attribution notifications (i.e. if the agent dies) to others. All received communication extends the graph with new agent node. The GNN used is GraphSAGE \cite{hamilton2017inductive}, with ten layers and a single-round communication, so many aggregations happen to enrich the final representation. GraphSAGE was designed for inductive learning and generalizes well to unseen nodes. The final representation is passed to the policy, which chooses the next node to visit. The main advantage is that the method shows resilience against noisy communication and agent attributions, compared to other patrolling algorithms.

Graph Proximal Policy Optimization (GPPO) and Heterogeneous GPPO (HetGPPO) \cite{bettini2023heterogeneous} are two models of communication extending IPPO \cite{de2020independent} with GNNs. GPPO uses parameter sharing while HetGPPO uses independent parameters. In detail, the local sub-graph is built in a predefined manner: bidirectional edges appear if nodes are in range of communication. Then, any MPNN (Eq. \ref{eq:MPNN}) process only once to aggregate received messages. Finally, $[o_i^t \oplus h_i^{t,1}]$ are fed into an MLP, and the output serves both policy and critic. The authors argue that HetGPPO leads to better heterogeneous behaviors, while being more resilient in noisy environments.

Heterogeneous Policy Networks (HetNet) studies communication between physically heterogeneous agents \cite{seraj2022learning}. The method posits that agents needs to learn different type of communication, depending on the class of agents. The graph is built in a predefined manner, like GPPO. Then, a custom MPNN called HetGAT learns different sets of weights for each class received messages. 
\MM{Concretely, if agent $i$ receives a message $x_j^t$ from an agent of a different class, it uses a dedicated attention weight $\alpha^{j2i}$. If it receives a message $x_k^t$ from an agent of the same class, it instead uses the attention weight $\alpha^{i2i}$.}
Moreover, HetGAT accounts for bandwidth by learning to sample fixed bit size messages. HetNet stacks several HetGAT layers, with \textit{multi-round communication} at each layer (cf. lines 12-14 of Algo. \ref{alg:comm}). Extending MAPPO, the final representation serves the policy during training and execution. For the centralized critic, the method uses a proxy as DICG to learn with a HetGAT on a complete graph. This method uses both a proxy and distributed communication (cf. lines 2 and 5 of Algo. \ref{alg:comm}), but the proxy is used only during the centralized training.

\subsection{Communication constraints}

\MM{To deploy MADRL with communication in real-world applications}, several \LM{other} constraints come into play, \LM{in particular concerning the communication. Indeed, in realistic scenarios, communication is subject to constraints and associated costs. Table \ref{tab:comm_constraints} provides a summary of the communication constraints evaluated during execution and/or accounted for in state-of-the-art methods. The first constraint relates to connectivity, characterized by a limited communication range (CR) among agents. While this constraint is not compatible with proxy-based architectures, it is naturally incorporated in distributed communication methods, especially, multi-round communication provides a solution to overcome this limitation. Another constraint is the scalability of the method in terms of number of agents.} Scaling to larger number of agents is easier in distributed communication, while a proxy communication suffers from greater complexity due to the size of the joint observation space. \LM{Another important aspect of realistic communication is the limited bandwidth (LB), which requires to optimally encode information when communication capacity is limited.} Few works begin to focus on this, as does HetNet \cite{seraj2022learning} or Bandwidth-constrained Variational Message Encoding (BVME) \cite{duan2025bandwidth} very recently. Moreover, noisy messages (NM) and communication loss (CL) are yet to be fully taken into account. Few methods are resilient to noisy messages \cite{bettini2023heterogeneous,goeckner2024graph}.

\begin{table}[h!t]
	\centering
	\caption{Tested communication constraints during execution}
	\label{tab:comm_constraints}
	\rowcolors{2}{gray!15}{white}
	\begin{tabular}{lccccc}
		\toprule
		Methods & CR & $\max(n)$ & LB &  NM & CL \\
		\midrule
		GA-Comm \cite{liu2020multi}&  & 20 &  &  & \\
		MAGIC \cite{niu2021multi}&  & 20 &  &  & \\
		GACG \cite{duan2024group}&  & 10 &  &  & \\
		LTS-CG \cite{duan2024inferring}&  & 27 &  &  & \\
		\midrule
		DGN \cite{Jiang2020Graph}& $\boldsymbol{\times}$ & 60 &  &  & \\
		LSC \cite{liu2023deep}& $\boldsymbol{\times}$ & 60 &  &  & \\
		MAGE-X \cite{yang2023learning}& $\boldsymbol{\times}$ & 50 &  &  & \\
		MAGEC \cite{goeckner2024graph}& $\boldsymbol{\times}$ & 6 &  & $\boldsymbol{\times}$ & $\boldsymbol{\times}$ \\
		GPPO \cite{bettini2023heterogeneous}& $\boldsymbol{\times}$ & 2 &  &  & \\
		HetGPPO \cite{bettini2023heterogeneous}& $\boldsymbol{\times}$ & 2 &  & $\boldsymbol{\times}$ & \\
		HetNet \cite{seraj2022learning}& $\boldsymbol{\times}$ & 10 & $\boldsymbol{\times}$ &  & \\
		\bottomrule
	\end{tabular}
\end{table}


To deploy communicating learning agents in real-world environment, integrating directly these constraints into the learning process of communication \LM{will enable the handling of more complex and realistic scenarios.}

\section{Conclusion} \label{sec:ccl}
Communication has emerged as a major solution to overcome partial observability and non-stationarity problems of Multi-Agent Deep Reinforcement Learning (MADRL) algorithms. Nevertheless, \MM{communication methods must address several key design challenges to be effective: determining the content of transmitted messages, selecting appropriate recipients, interpreting and combining received information, and leveraging communication-aware representations for decision-making. Our survey highlights the use of Graph Neural Networks (GNNs) in the Multi-Agent Deep Reinforcement Learning (MADRL) literature as a principled approach to partially address these four challenges.}

We propose a generic algorithm that generalizes the GNN-based communication processes, which presents how GNNs handle communication challenges. Our algorithm captures the diversity in methods using GNNs and permits a comparative study of existing approaches, emphasizing common and different structural elements and limitations.
\MM{We have classified existing methods from two complementary perspectives: proxy communication and distributed communication. The proxy assumption enables the construction of a global graph, facilitating stronger coordination among agents. However, it relies on the assumption of perfect communication. In contrast, distributed communication supports local coordination within communication range, reflecting common practical constraints.} In addition, we show the limitations of GNN-based communication methods in communication-constrained environments.

Lastly, we observe that taking into account the temporal information and the structural information is a key point for better communication between agents. Encoding messages often uses an RNN to aggregate observations from the past, and its help to converge in a partially observable environment. However, the correlation of temporal and structural information, as LTS-CG leverages \cite{duan2024inferring}, can be a potential research direction.

The validation of the proposed generalization is intended to rely on the design of a unified framework enabling systematic comparison, reproducibility and empirical verification across existing and future methods.

\MM{Finally, future research should more explicitly address realistic constraints, as summarized in Table \ref{tab:comm_constraints}, to bridge the gap between theoretical communication models and practical applications of multi-agent systems.}

\bibliographystyle{plain}
\bibliography{biblio-ch-pfia}

\end{document}